\documentclass[runningheads]{llncs}
\usepackage{graphicx}
\usepackage{amsmath,amssymb} 
\usepackage{color}
\usepackage[width=122mm,left=12mm,paperwidth=146mm,height=193mm,top=12mm,paperheight=217mm]{geometry}

\usepackage{times}
\usepackage{epsfig}
\usepackage[pagebackref=true,breaklinks=true,letterpaper=true,colorlinks,bookmarks=false]{hyperref}

\DeclareMathOperator*{\argmin}{arg\,min}
\DeclareMathOperator*{\argmax}{arg\,max}
\DeclareMathOperator*{\vis}{vis}

\newcommand{\red}[1]{\textcolor{black}{#1}}

\begin{document}

\pagestyle{headings}
\mainmatter
\def\ECCV16SubNumber{50}  

\title{Temporally Consistent Motion Segmentation from RGB-D Video}

\titlerunning{Temporally Consistent Motion Segmentation from RGB-D Video}

\authorrunning{P. Bertholet, A. Ichim, M. Zwicker}

\author{P. Bertholet\textsuperscript{1}, A. Ichim\textsuperscript{2}, M. Zwicker\textsuperscript{1}}
\institute{\textsuperscript{1}University of Bern, Switzerland \\
\textsuperscript{2}\'{E}cole polytechnique f\'{e}d\'{e}rale de Lausanne, Switzerland}

\maketitle

\begin{abstract}
   We present a method for temporally consistent motion segmentation from RGB-D videos assuming a piecewise rigid motion model. We formulate global energies over entire RGB-D sequences in terms of the segmentation of each frame into a number of objects, and the rigid motion of each object through the sequence. We develop a novel initialization procedure that clusters feature tracks obtained from the RGB data by leveraging the depth information. We minimize the energy using a coordinate descent approach that includes novel techniques to assemble object motion hypotheses. A main benefit of our approach is that it enables us to fuse consistently labeled object segments from all RGB-D frames of an input sequence into individual 3D object reconstructions.

\end{abstract}

\section{Introduction}
Leveraging motion for object segmentation in videos is a well studied problem. In addition, with RGB-D sensors it has become possible to exploit not only RGB color data, but also depth information to solve the segmentation problem. The goal of our approach is to allow a user, or a robotic device, to move objects in a scene while recording RGB-D video, and to segment objects based on their motion. We assume a piecewise rigid motion model, and do not constrain camera movement. This scenario has applications, for example, in robotics, where a robotic device could manipulate the scene to enhance scene understanding~\cite{Ma2015SLM}. Another application scenario is 3D scene acquisition, where a user would be enabled to physically interact with the scene by moving objects around as the scene is being scanned. The system would then segment and reconstruct individual objects, instead of returning a monolithic block of geometry. KinectFusion-type techniques enable similar functionality, but with the restriction that a full scene reconstruction needs to be available before segmentation can start~\cite{Izadi2011KRR}. In contrast, we do not require a complete scan of an entirely static scene.

We formulate joint segmentation and piecewise rigid motion estimation as an energy minimization problem. \red{In contrast to previous approaches our energy encompasses the entire RGB-D sequence, and we optimize both motion and segmentation globally instead of considering only frame pairs.} 
This allows us to consistently segment objects by assigning them unique labels over complete sequences. Our approach includes a novel initialization approach based on clustering feature trajectories by exploiting depth information. We perform energy minimization using a coordinate descent technique, where we iteratively update object segmentation and motion hypotheses. A key contribution is a novel approach to recombine previous motion hypotheses by cutting and re-concatenating them in time to obtain temporally consistent object motions. To avoid bad local minima, we develop a novel initialization strategy that clusters feature trajectories extracted from the RGB frames by exploiting the  depth information from the RGB-D sensor. Finally, we demonstrate that we can fuse object segments obtained from all input frames into consistent reconstructions of individual 3D objects.

\begin{figure*}[t]
\includegraphics[width=\textwidth]{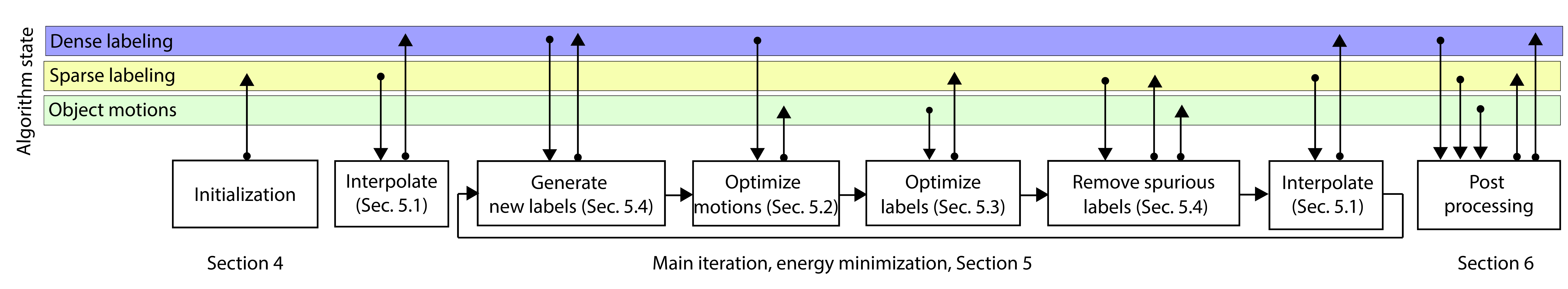}%
\caption{Overview: After an initialization, we perform iterative energy minimization over object segmentation and object motions. For computational efficiency we operate on two different scene representations, a sparse and a dense one, and include an interpolation step to go from sparse to dense. Vertical arrows indicate read and write operations into the scene representations.
}
\label{fig:pipeline}%
\end{figure*}

\section{Previous Work}


\paragraph{Motion Segmentation from RGB Video.} Motion segmentation from video is a classical topic in computer vision, and a full review is beyond the scope of this paper. We are inspired by the state of the art approach by Ochs et al.~\cite{Ochs2014SMO}. They observe that motion is exploited most effectively by considering it over larger time windows, for example by tracking feature point trajectories. Then, dense segmentation is obtained in a second step. They point out the advantage of obtaining consistent segmentations over entire video sequences, which is also a goal in our approach. Our initialization step follows a similar pattern as Ochs et al., where we track and cluster feature trajectories. However, a key distinction is that we exploit the depth data to aid clustering. Recently, learning based approaches have also become popular for motion segmentation~\cite{Fragkiadaki2015LSM}, and it may be interesting in the future to apply such techniques also to RGB-D data.

\paragraph{Motion Segmentation from RGB-D Data.} A number of techniques have been quite successful in segmenting objects from pairs of RGB-D frames. Our work is most related to the recent approach of St\"uckler et al.~\cite{Stueckler2015EDR} who also use a piecewise rigid motion model and perform energy minimization to recover object segmentation and motion simultaneously. Similar to our approach, they use a coordinate descent strategy for energy minimization and graph cuts to update the segmentation. Earlier work includes the approach by Ven et al.~\cite{Ven2010IPM}, who also jointly solve for segmentation and motion by formulating a CRF and using belief propagation. Both techniques, however, are limited to pairs of RGB-D frames. The main difference to our technique is that we solve globally over an entire RGB-D sequence, which allows us to consistently label segments, track partial objects, and accumulate data over time. 

Our problem is similar to other techniques that leverage entire RGB-D sequences to segment objects based on their motion, and to fuse partial objects over all frames into more complete 3D reconstructions. The original KinectFusion system~\cite{Izadi2011KRR} can segment moving objects after a complete scan of a static scene has been obtained. Perera et al.~\cite{Perera2015MST} improve on this by segmenting objects based on incremental motion, whereas KinectFusion requires objects to move completely outside their originally occupied volume in the static scene. As a crucial difference to our approach, both approaches rely on a complete 3D reconstruction of a static version of the scene that needs to be acquired first, before object segmentation can be performed. 

The goal of Ma and Sibley's work~\cite{Ma2014UDO} is the most similar to ours, as they discover, track and reconstruct objects from RGB-D videos based on piecewise rigid motion. A key difference is that they use an incremental approach as they move forward over time to discover new objects, by detecting parts of the scene that cannot be tracked by the dominant camera motion. This means that groups of objects that initially exhibit the same motion (for example one object moving on top of another), but later split and move along different trajectories, cannot be consistently identified and separated over the entire sequence. In contrast, we optimize jointly over segmentation and motion, taking into account entire RGB-D sequences, instead of incremental segmentation followed by tracking. This allows to successfully resolve such challenging scenarios.

\paragraph{RGB-D Scene Flow.} Our problem is also related to the problem of obtaining 3D scene flow, that is, frame-to-frame 3D flow vectors, from RGB-D data. For example, Herbst et al.~\cite{Herbst2013DME} generalize two-frame variational 2D flow to 3D, and apply it for rigid motion segmentation. Quigoro et al.~\cite{Quiroga2014DSR} model the motion as a field of twists and they encourage piecewise rigid body motions. They do not address segmentation, and their method processes pairs of RGB-D frames separately. Sun et al.~\cite{Sun2015LSF} also address scene flow, but they formulate an energy over several frames in terms of scene segmentation and flow. While they can deal with several moving objects, their segmentation is separating depth layers. They also show results only for short sequences of less than ten frames. 
\red{Jaimez et al. ~\cite{jaimez2015motion} leverage a soft piecewise rigidity assumption and jointly optimize for segmentation and motion to extract high quality scene flow and segmentations for pairs of RGB-D frames.}
In contrast, our goal is to separate objects only based on their individual motion, and label the segmented objects consistently over time. We perform energy minimization on video segments instead of frame pairs, which also allows us to fuse data over time into 3D object reconstructions \red{and reason explicitly about occlusion}.

\section{Overview}

Given a sequence of RGB-D images as an input, our goal is to assign an object label to each RGB-D pixel (that is, each acquired scene point), and to track the motion of all objects through the sequence. We assume a piecewise rigid motion model, and we define objects as groups of scene points that exhibit the same motion trajectories through the entire sequence. We do not assume any a priori knowledge about object geometry or appearance, or the number of objects, and camera motion is unconstrained. 

Figure~\ref{fig:pipeline} shows an overview of our approach. In the main iteration, we solve an energy minimization problem, where the energy is defined as a function of pixel labels, and per label object motions, that is, sequences of rigid transformations. We describe the energy minimization in detail in Section~\ref{sec:energyminimization}. For computational efficiency, energy minimization operates on two different scene representations, a sparse and a dense one. We include an interpolation step to go from sparse to dense. Since the energy is non-linear and has many local minima, it is important to start coordinate descent with a good initialization, as described next.


\section{Initialization}
\label{sec:initialization}

The goal of the initialization step is to group a set of sparse scene point trajectories into separate clusters, each cluster representing an object hypothesis and its rigid motion. We obtain the sparse scene point trajectories using 2D optical flow similar to the work by Ochs et al.~\cite{Ochs2014SMO}. Each trajectory spans a temporal subwindow of frames from the input sequence. The motivation to start with longer term trajectories, as opposed to pairwise processing of frames, is that longer trajectories that overlap in time enable the algorithm to share information globally over an entire input sequence, for example, to propagate segmentations to instants where objects are static. 

We denote a point trajectory obtained via 2D optical flow by $t = (p^t_k, p^t_{k+1}, \dots, p^t_l)$. The trajectory $t$ is supported through consecutive frames $k\dots l$, and $p^t_k, \dots, p^t_l$ are the 3D positions along the track. We denote the set of all trajectories by $T$. A key idea in our approach is to leverage the 3D information to cluster the tracks. Note that each cluster (that is, each subset of $T$) directly implies a sequence of rigid transformations that best aligns all points $(p^t_k, p^t_{k+1}, \dots, p^t_l)$ in each track $t$ in the cluster. Hence, we can cluster immediately by minimizing the total alignment error over all clusters.

We implement this idea using a soft clustering approach. Each trajectory $t$ has a weight vector $w_h(t)$, where $h \in 1\dots N$, $h$ represents a cluster, and $N$ is the number of clusters. We restrict the weights to be positive and sum up to one. Intuitively, $w_h(t)$ represents the probability that trajectory $t$ belongs to cluster $h$. Similar as in hard clustering, all the trajectory weights $\{w_h(t)\}_{t \in T}$ for a single cluster $h$ directly imply a sequence of rigid transformations that minimize a weighted alignment error. Hence we can write the rigid transformation between arbitrary frames $i$ and $j$ for cluster $h$ as a non-linear function of its weights $\{w_h(t)\}_{t \in T}$, denoted $A^{i \rightarrow j}_h(\{w_h(t)\}_{t \in T})$. The total alignment error over all trajectories and clusters then can be seen as a function of the weights for each trajectory and cluster $\{w_h(t)\}_{t \in T, h=1\dots N}$, which we denote by $w$ for simplicity,
\begin{align}
\label{eq:initializationenergy}
&E_{\mathit{init}}(w) = \sum_{h=1}^N \sum_{t\in T} w_h(t) \sum_{\{ k | k \in t\}} d\left(A^{t_{\mathit{seed}} \rightarrow k}_h(\{w_h(t)\}_{t \in T})p^t_{t_\mathit{seed}},p_k^t\right). 
\end{align}
The innermost sum is over all frames of trajectory $t$, suggestively denoted by $\{ k | k \in t\}$. It measures the alignment error for trajectory $t$ under the hypothesis that it belongs to cluster $h$ by transforming a selected point $p^t_{t_\mathit{seed}}$ to all other frames. We set $t_\mathit{seed}$ to the frame in the middle of the trajectory. Finally, $d(\cdot,\cdot)$ is is the point-to-plane distance.

We minimize this energy using gradient descent with approximate gradients. For efficiency, we consider the alignments $A^{t_{\mathit{seed}} \rightarrow k}_h$ to be constant in each gradient descent step. Hence the approximate partial derivatives with respect to the weights are
\begin{align}
\frac{\partial}{\partial w_h(t)} E_{\mathit{init}}(w) \approx 
\sum_{\{ k | k \in t\}} d\left(A^{t_{\mathit{seed}} \rightarrow k}_h p^t_{t_\mathit{seed}},p_k^t\right), 
\end{align}
and they form our approximate gradient $\nabla E_{\mathit{init}}$. 

The gradient descent step $\Delta w$ needs to maintain the constraint that the weights are positive and form a partition of unity. This is done by projecting $\Delta w$ onto the corresponding subspaces. In order to keep the local alignment constancy assumption we scale the $\Delta w$ to a fixed norm $\epsilon$.
After each weight update we solve for new alignments $A^{t_{\mathit{seed}} \rightarrow k}_h$ by minimizing Equation~\ref{eq:initializationenergy} using the updated weights. This minimization is performed using a Levenberg-Marquardt approach.



\section{Energy Minimization}
\label{sec:energyminimization}

Our energy is defined as a function of the labeling of scene points with object hypotheses, and the sequences of transformations for each object hypothesis. The energy consists of a spatial and a temporal smoothness term, and a data term. We perform energy minimization using a coordinate descent approach, where we update transformation sequences and label assignments in an interleaved manner (see Figure~\ref{fig:pipeline}). For computational efficiency we use slightly different data terms in these two steps and do not always use the dense data. When updating motion sequences (Section~\ref{sec:bundleadjustment}) we use a data term based on the densely sampled input scene points. In the optimization of the label assignments (Section~\ref{sec:labelassignments}), however, we use a sparsely sampled representation of the input. We use a simple interpolation procedure to upsample the label assignments back to the input data (Section~\ref{sec:interpolation}). Finally, we include heuristic strategies to add or remove labels during the iteration to avoid getting stuck in local minima (Section~\ref{sec:labelgeneration}). 


\subsection{Interpolating Labels} 
\label{sec:interpolation}

We upsample labels on a sparse set of scene points to the input scene points using a simple interpolation approach. This step is necessary after initialization as well as after the sparse label assignment step in each iteration (Figure~\ref{fig:pipeline}). We interpolate in 3D using a local weighted averaging of labels based on the Euclidean distances between dense interpolation target points and the sparse labeled samples. After interpolation we obtain weights $w_h(p)$ for each label $h$ and scenepoint $p$. Note that these weights are continuous (as opposed to binary), because of the weighted averaging.

\subsection{Optimizing the Object Motions}
\label{sec:bundleadjustment}

Given a labeling of all input scene points with an object hypothesis $h \in \{1\dots N\}$, our goal in this step is to compute transformations $A_h^{i \rightarrow j}$ for all labels $h$, which best align their data between a certain set of pairs of frames $(i,j)$. 

\paragraph{Alignment Error.} Since the motion is optimized for each label independently we drop $h$ for simplicity. We write $A^{i \rightarrow j}$ as a transformation of frame $i$ to a reference coordinate system, followed by the transformation to frame $j$, that is, $T_j T_i^{-1}$. We assume the point correspondences between frames $i$ and $j$ have been computed where the $k^{th}$ correspondence is denoted as $(p^i_k, p^j_k)$. We weight the correspondences based on their relevance for the current label by $w_{\max} = \max \{w_h(p_k^i), w_h(p_k^j)\}$. Our alignment error is a sum of point-to-plane and point-to-point distances,
%
\begin{align}
\label{eq:alignmenterror}
E = \sum_{(i,j)} \sum_k w_{\max} &\left( \left<T_jT_i^{-1}p_k^i - p_k^j, n_k^j\right>^2 + \alpha \| T_jT_i^{-1}p_k^i - p_k^j \| \right),
\end{align}
where $\alpha$ balances the two error measures, and $n_k^j$ is the normal of vertex $p_k^j$.

\paragraph{ICP Iteration.} We solve for the rigid transformations $T_i$, $T_j$ by alternating between updating point correspondences and minimizing the alignment error from Equation~\ref{eq:alignmenterror} in a Levenberg-Marquardt algorithm. For faster convergence, we solve for the transformations in a coarse-to-fine fashion by subsampling the point clouds hierarchically and using clouds of increasing density as the iteration proceeds.


\paragraph{Selecting Frame Pairs.} 
The simplest strategy is to include only frame pairs $(i,j=i+1)$ in Equation~\ref{eq:alignmenterror}. 
This may be sufficient for simple scenes with large objects and slow motion. However, it suffers from drift. We enhance the incremental approach with a loop closure strategy to avoid drift similar to Zollh\"ofer et al.~\cite{Zollhofer2015SRV}. The idea is to detect non-neighboring frames that could be aligned directly, and add a sparse set of such pairs to the alignment error in Equation~\ref{eq:alignmenterror}. We use the following heuristics to find eligible pairs:
\begin{itemize}
\item The centroid of the observed portion of the object in frame $i$ lies
in the view frustum when mapped to frame $j$, and vice versa. 
\item The viewing direction onto the object, approximated by the direction from the camera to the centroid, should be similar in both frames. We tolerate a maximum deviation of $45$ degrees. 
\item The distance of the centroids to the camera is similar in both frames. Currently we tolerate a maximum factor of $2$. 
\end{itemize}
The first two criteria are to check that similar parts of the object are visible in both frames, and they are seen from similar directions. The third one ensures that the sampling density does not differ too much. Initializing a set $S$ with the adjacent frame constraints $(i,i+1)$, we greedily extend it with a given number of additional pairs $(k,l)$ from the eligible set. We iteratively select  and add new pairs from the eligible pairs such that they as distant as possible from the already selected ones:
\begin{align}
S\leftarrow S \cup \argmax_{eligible\;(k,l)} \left( \min_{(i,j) \in S} |k-i|+|j-l| \right)
\end{align}
%
Overall for our ICP variant with loop closures we first solve for alignments only with the neighboring frame pairs $(i,i+1)$, taking identity transformations as initial guesses of the alignment between adjacent frames. We then use these alignments to determine and select additional eligible loop closure constraints and do a second ICP iteration with the extended set of frame pairs.

\subsection{Optimizing the Labels}
\label{sec:labelassignments}

The input to this step is a set of  motion hypotheses $h\in\{1,...,N\}$, that is, a set of transformation sequences $A_h^{i \rightarrow i+1}$ over all frames $i$, which describe how a scene point could move through the entire sequence. The output is a labeling of scene points with one of the motion hypotheses $h$. The idea is to assign the motion to each point that best fits the observed data and yields a spatio-temporally consistent labeling.

In this step we operate on a sparse set of scene points $P$, which we obtain by spatially downsampling the dense input scene points in each frame separately. Each $p\in P$ has a seed frame $s_p$ where the point was sampled.
Denote the labeling of the sparse points $p \in P$ by the map $L : p \mapsto h(p) \in \{1,\dots, N\}$, where $N$ is the number of labels. 
We find the labeling by minimizing an energy consisting of a data and a smoothness term,
\begin{align}
\label{eq:energy}
\argmin_L \sum_{p \in P} -\log(\mathcal L_h(p|Data)) + \sum_{(p,q) \in P \times P} V_h(p,q),
\end{align}
%
\red{where $\mathcal L_h(p| Data)$ measures the likelihood that $p$ moves according to the attributed motion $h(p)$ given the dense input $Data$. In addition, $V_h(p,q)$ is a spatio-temporal smoothness term. We minimize the energy using graph cuts and $\alpha-\beta$ swaps ~\cite{boykov2001fast}.}

\paragraph{Data Term.} 
\red{We formulate the likelihood $\mathcal L_h(p| Data)$ for a motion hypothesis $h(p)$ to be related inversely to a distance $d_h(p, Data)$. This distance measures how well mapping $p$ to all frames according to $h(p)$ matches the dense observed data, as described in more detail below. For each point $p$, we normalize the distances $d_h(p, Data)$ over all possible assignments $h(p)$ to sum to one. Then, for each assignment $h(p)$, we map its normalized distance to the range $[0,1]$, and assign one minus this value to $\mathcal L_h(p| Data)$.}
\red{The advantage of this procedure is that the resulting likelihoods only depend on relative magnitudes of observed distances; absolute distances would decrease throughout the iteration and might also vary spatially with the sampling density and noise level of the depth sensor.}

\red{ We design the distance $d_h$ to be robust to outliers, to explicitly model and factor out occlusion, and to take into account that alignments might be corrupted by drift. Let $p^f$ be the location of $p$ mapped to frame $f$ using the motion of its current label. More precisely, $p^f = A_{h(p)}^{s_p \rightarrow f} p$. The trajectory of $p$ over all frames is $\{p^f\}$.} Denoting the nearest neighbor of $p$ in a frame $f$ by $\mathrm{NN}^f(p)$ and the clamped $L_2$ distance between a point and its neighbor by $d_{NN}^f(p) := ||p-NN^f(p)||^2_{clamped} $ we formulate it as
\begin{align}
\label{eq:distancemeasure}
&d_h(\{p^f\}, Data) = \frac{1}{\sum_f \vis(p^f)} \; \cdot \\
&\sum_f \vis(p^f)\left[ d_{\mathrm{NN}}^{f}(p^{f}) + \alpha\sum_{i=\pm1}d_{\mathrm{NN}}^{f+i}(A_{h(p)}^{f\rightarrow f+i}(\mathrm{NN}^{f}(p^{f})))\right].\nonumber
\end{align}
A key point is that all motion hypotheses may be contaminated by drift. Hence we also take the incremental error due to the transformation $A_{h(p)}^{f\rightarrow f\pm1}$ to neighboring frames into account and balance the terms, in our experiments with $\alpha = 0.5$. \red{If $p^f$ is further away from its neighbor than the clamping threshold, we set the incremental errors to the maximum.}

Occlusion is modeled explicitly by $\vis(p^f)$, which is a likelihood of point $p$ being visible in frame $f$. We formulate this as a product of the likelihood that $p^f$ is facing away from the camera and the likelihood of $p^f$ being occluded by observed data,
\begin{align}
&\vis(p^f) = \mathrm{clamp}_0^1\left(\frac{\left< p^f.n, p^f.v \right>}{0.3}\right) \cdot \begin{cases} 1 & \text{if $\pi(p^f)$ behind $p^f$} \\
 \frac{\sigma^2}{\sigma^2 + (p^f.z-\pi(p^f).z)^2} & \text{else,}
\end{cases}
\end{align}
where $p^f.n$ is the unit normal of $p^f$, $p^f.v$ is the unit direction connecting $p^f$ to the eye, $p^f.z$ is its depth, $\pi(p^f)$ is the perspective projection of $p^f$ onto the dense depth image of frame $f$, and $\sigma^2$ is an estimate for the variance of the sensor depth noise. In addition, the visibility $\vis(p^f)$ is set to zero if $p^f$ is outside the view frustum of the sensor, or the projection $\pi(p^f)$ is mapped to missing data.

Note that the complexity to compute the data term in Equation~\ref{eq:energy} is quadratic in the number of frames ($|P|$ is proportional to the number of frames), hence it becomes prohibitive for larger sets of frames. Therefore, we compute contributions to the error only on a subset of frames. The $k$ frames adjacent to the seed frame of a point are always evaluated, the frames further away are sampled inversely proportional to their distance to the seed frame. \red{This also effectively weights down the contribution of distant frames; through the choice of a heavy tailed distribution their contributions remain relevant.} Finally, a motion hypothesis may not be available for all frames. If $p^f$ is not defined because of this, we set all corresponding terms in Equation~\ref{eq:distancemeasure} to zero.

\paragraph{Smoothness Term.} The smoothness term is
\begin{align}
V_h(p,q) = 
-\log\left(1-\frac{\sigma^2}{\sigma^2+\|p-q\|^2} \right)
\end{align}
if $q$ is in a spatio-temporal neighborhood $\mathcal{N}(p)$ of $p$, and the labels differ, $h(p) \neq h(q)$. Otherwise the smoothness cost $V(p,q)$ is zero. The norm here is simply the squared Euclidean distance. The neighborhood $\mathcal{N}(p)$ includes the $k_1$ nearest neighbors in the seed frame of $p$, and the $k_2$ nearest neighbors in one frame before and after.

\subsection{Generating and Removing Labels}
\label{sec:labelgeneration}

\begin{figure}[t]
\begin{center}
\includegraphics[width=\columnwidth]{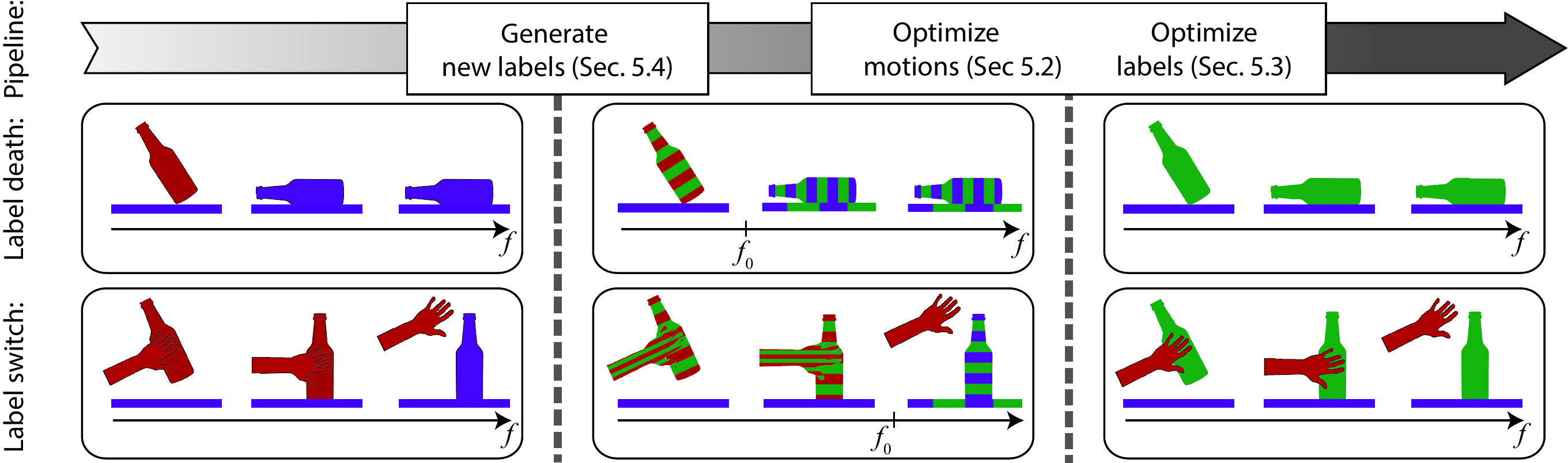}
\end{center}
\caption{We illustrate our heuristics to break out of local minima. Middle left: The bottle first has its own motion (tipping over), then remains static with the table. The red label assigned to the bottle ``dies''. Bottom left: The bottle first shares its motion with the hand, then remains static with the table. Its label switches from red (associated with the motion of the hand) to blue (motion of the table). These configurations are local minima in our optimization. Middle column: We resolve this by detecting label events (birth, death, switch) at frames $f_0$. We add a new label (green) to all red points up to the label event at $f_0$, and to all blue point starting from $f_0$ (the striped objects now have two labels). Right column: The subsequent motion and label optimization steps extract and assign an additional motion for the green label that resolves the mislabeling.}
\label{fig:localminimum}%
\end{figure}

Scenes where different objects exhibit different motions only in a part of the sequence, but move along the same trajectories otherwise, are challenging. As illustrated in Figure~\ref{fig:localminimum} (middle row, left), our iteration can get stuck in a local minimum where one of the objects gets merged with the other and its label disappears as they start moving in parallel. In the figure, the bottle tips over and remains static with the support surface after the fall, and the red label disappears. We term this \textit{``label death''}. Analogously, a label may emerge as two objects split and start moving independently (\textit{``label birth''}). Finally (Figure~\ref{fig:localminimum}, bottom left), a first object (the bottle) may first share its motion with a second one (the hand), and then with a third one (the support surface). Hence the first object (bottle) may first share its label with the second one (hand), and then switch to the third one (support surface). We call this \textit{``label switch''}. These three cases are local minima and fixpoints of our iteration. In the situations in Figure~\ref{fig:localminimum}, left column, the motions are optimal for the label assignments and the labels are optimal for the motions. 



We use a heuristic to break out of these local minima by introducing a new label, which is illustrated in green color in Figure~\ref{fig:localminimum} (middle column), consisting of a combination of the two previous labels (blue and red). The key challenge for this heuristics is to detect a frame $f_0$ where a label event (death, birth, or switch) occurs, as described below. Then we add the new label (green) to the points labeled red before $f_0$ and the points labeled blue after $f_0$ (because our labels are continuous weights at this point, each point may have several labels with non-zero weights; this is illustrated by the stripe pattern in the figure). In the next motion optimization step (Section~\ref{sec:bundleadjustment}), the green label will lead to the correct motion of the previously mis-labeled bottle, and the subsequent label optimization step (Section~\ref{sec:labelassignments}) will correct its label assignments, yielding the configuration in the right column in Figure~\ref{fig:localminimum}. In general, there are more than two labels in a scene, hence we also need to determine which pair of labels is involved in a label event, which we describe next.


\paragraph{Detecting Label Events.} We detect label death if a label is used on less than $0.5\%$ of all pixels in a frame, and label birth if a label assignment increases from below $0.5\%$ to above $0.5\%$ from one frame to the next. To detect label switches, we analyze for each pair of labels how many pixels change from the first to the second label. As we encode labels with continuous weights, we want to measure how much mass is transferred between any two labels. The intuition is that local extrema in mass transfer correspond to label switches. We represent mass transfer in an $N \times N$ matrix $M^f$ for each frame $f$, where $N$ is the number of labels. Note that we only capture the positive transfer of mass between labels.  

More precisely, for any pixel $p$ in a frame $f$ with weights $w(p) = (w_1(p),...,w_N(p))^\intercal$ we estimate its weights $w'(p)$ in the next frame by interpolating the weights of its closest neighbors in the next frame, and we compute their difference $\Delta w = w' - w$. We then estimate the weight transfer matrix $M(p)$ for this single pixel by distributing the weight losses $L(p) = -\min(\Delta w,\textbf{0})$ proportionally on the weight gains $G(p) = \max(\Delta w,\textbf{0})$, that is $M(p) = \frac{1}{\sum G(p)} L(p) \cdot G(p)^\intercal$. The weight transfer matrix for a frame is then given by summing over all pixels, $M^f = \sum_{p\in frame\; f} M(p)$.

\begin{figure}[t]
\begin{center}
\includegraphics[width=\columnwidth]{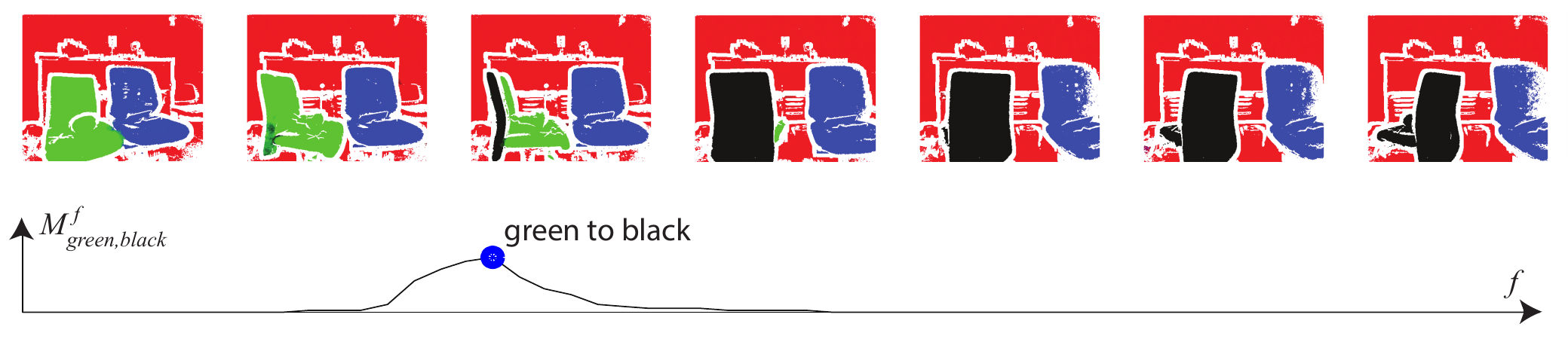}%
\end{center}
\caption{Visualization of the detection of a label switch. The graph plots the entries $M^f_{green,black}$ of the mass transfer matrices $M^f$ corresponding to weight transfer from green to black as a function of the frames $f$.}
\label{fig:measuringlabelgeneration}%
\end{figure}

Finally, we detect label switch events as the local maxima in the temporal stack of the matrices $M^f$; for example the most prominent switch from some label $\hat{i}$ to some $\hat{j}$ happening in some frame $\hat{f}$ is given by
$(\hat{i}, \hat{j},\hat{f}) = \argmax_{i, j ,f} M^f_{ij}$.
We select a fixed number of largest local maxima as label switches. To do local maximum suppression we also apply a temporal box filter to the matrix stack for each matrix entry.



\paragraph{Removing Spurious Labels.} At the end of each energy minimization step we finally remove spurious labels that are assigned to less than $0.5\%$ of all pixels over all frames (see Figure~\ref{fig:pipeline}).

\section{Post-Processing}

\paragraph{Fusing Labels.} Our outputs of the energy minimization sometimes suffer from oversegmentation because we use a relatively weak \red{spatial} smoothness term to make our approach sensitive to alignment errors, which is important to detect small objects. \red{Regions that allow multiple rigid alignments (planar, spherical or conical surfaces) slightly favor motions that maximize their visibility; if they are only weakly connected to the rest of the scene they are oversegmented.} In post-processing, we fuse labels if doing so does not increase the data term (alignment errors) significantly. For each pair of labels $i$ and $j$, we compute the data term in Equation~\ref{eq:energy} given by fusing label $i$ into label $j$. This operation implies that we use the motion of label $j$ for the fused object, hence it is not symmetric in general. We accept the fused label if the data cost does not increase more than $2\%$ over the one of label $j$.


\begin{figure}[t]%
\includegraphics[width=\columnwidth]{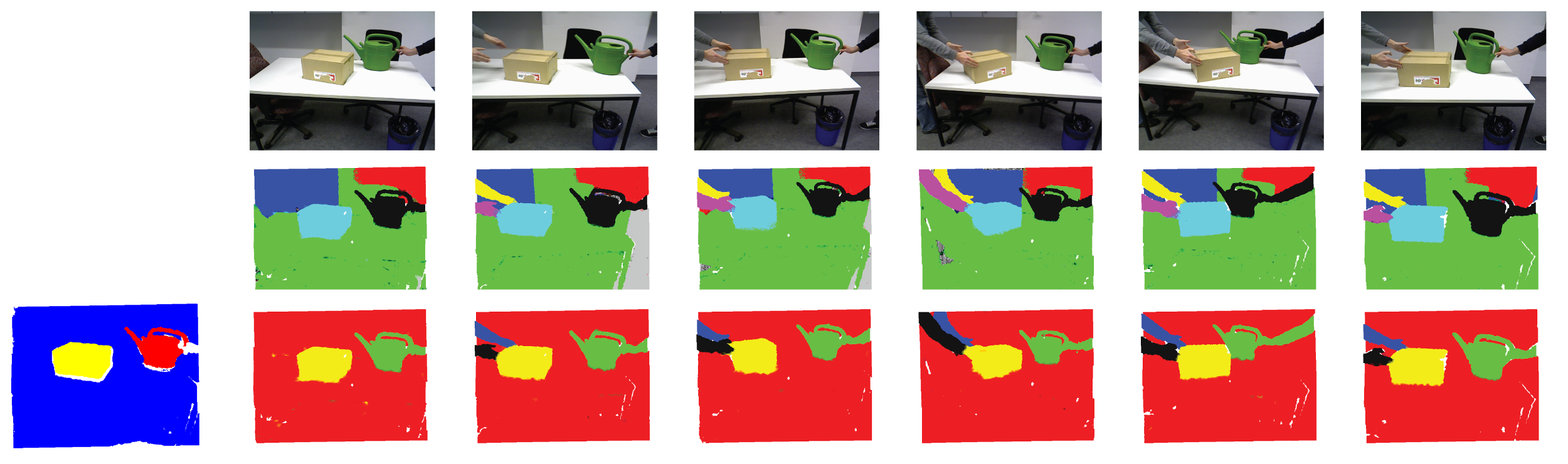}%
\caption{Results on a subsequence of the wateringcan\_box scene of St\"uckler and Behnke~\cite{Stueckler2015EDR}. Left: ground truth provided by St\"uckler and Behnke for the first depicted frame, note that non-rigid objects (like arms) were labeled as ``don't care'' (white) by them.  From top to bottom: input RGB, our output before post-processing, and after post-processing. Due to large noise levels and poor geometry in the background (set of parallel planes) as well as little spatial connectivity between the background and the foreground the smoothness term of the graph cut alone cannot prevent oversegmentation, but our post-processing step successfully fuses the  correct labels.}%
\label{fig:bonn_wateringcan_segmentation}%
\end{figure}

\section{Implementation and Results}

We implemented our approach in C++ and run all steps on the CPU. Our unoptimized code requires between fifteen minutes and about two hours per iteration of our energy minimization for the scenes shown below. We always stop after seven iterations. \red{In our experiments we only processed sequences of up to 220 frames such that processing times are under 24 hours per sequence.} 

To demonstrate our method we show our segmentation results, as well as \red{accumulated point clouds} and TSDF (truncated signed distance field) reconstructions of identified rigid objects, by directly using the segmentation masks and alignments obtained by out method. \red{We include the full results and the segmentation after each iteration in the supplementary material to further document the convergence of our method.} For volumetric fusion and cube marching we used the  InfiniTam library~\cite{InfiniTAM_ISMAR_2015}. Note that we did \textit{not} utilize any of the other features provided by Infinitam (like tracking) to improve our results.

We first show results of our approach on data sets provided by St\"uckler and Behnke~\cite{Stueckler2015EDR}. Figure~\ref{fig:bonn_wateringcan_segmentation} shows frames from a subsequence of their wateringcan\_box scene. We temporally subsampled their data to about $10$ frames per second, and process a subsequence of $60$ frames. The figure shows our segmentation on a selection of six frames. It demonstrates that we obtain a consistent labeling over time. We cannot separate the hand and the watering can here, since the hand holds on to the can through the entire sequence. Note that we do not perform any preprocessing of the data, such as manually labeling the hands using ``don't care'' labels, as \cite{Stueckler2015EDR} do. Figure~\ref{fig:bonn_wateringcan_segmentation} also shows the ground truth segmentation for the first frame, as provided by St\"uckler and Behnke. Note that in the ground truth hands and arms were labeled  white as ``don't care''. Finally, Figure~\ref{fig:wateringcan_reconstructions} left column shows reconstructions through volumetric fusion of two objects in this sequence \red{and an accumulated point cloud created by mapping the data of all $60$ frames to the central frame.}

\begin{figure}[t]
\includegraphics[width=\columnwidth]{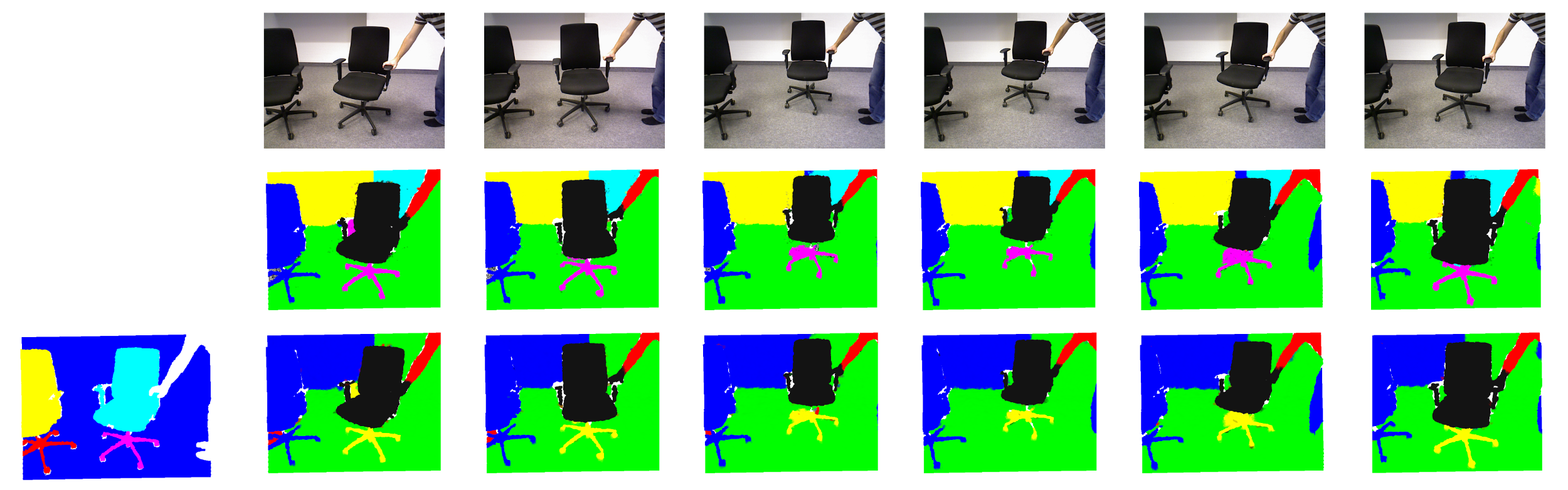}%
\caption{Results from a subsequence of the chair scene by St\"uckler and Behnke. From top: RGB data, our output before post-processing, and after post-processing. Our method was not able to achieve perfect results on the chair sequence, since the chair on the left is not moved during the subsequence we processed. On the other hand the green segment is geometrically very poor: it consists of the back wall and the floor, hence finding alignments is challenging.}%
\label{fig:bonn_chair_segmentation}%
\end{figure}
\begin{figure}%
\begin{center}
\includegraphics[width=0.9\columnwidth]{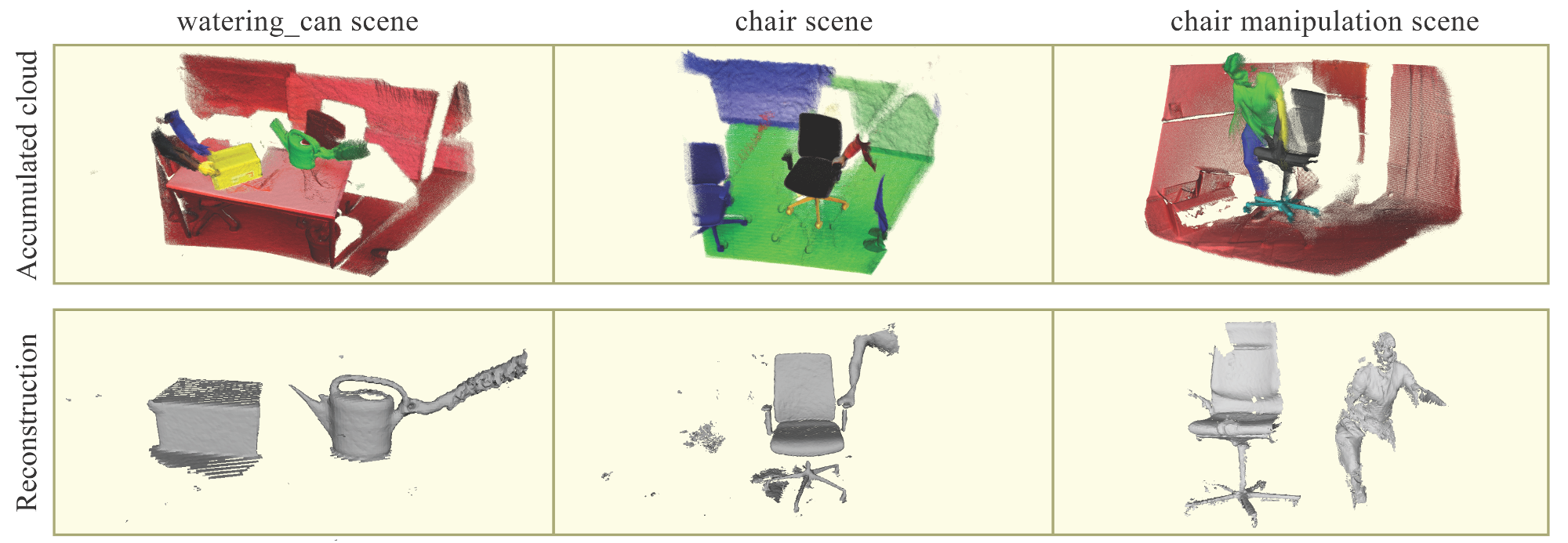}%
\end{center}
\caption{From left to right: Accumulated point clouds and selected reconstructions for the wateringcan\_box sequence (Figure~\ref{fig:bonn_wateringcan_segmentation}), the chair sequence (Figure~\ref{fig:bonn_chair_segmentation}) and our chair manipulation sequence (Figure~\ref{fig:own_chair_segmentation_convergence}). The accumulated point clouds were created by mapping all observed data points to the central frame.}%
\label{fig:bonn_chair_reconstruction}%
\label{fig:wateringcan_reconstructions}%
\label{fig:ownchair_reconstructions}%
\end{figure}

Figure~\ref{fig:bonn_chair_segmentation} shows our results on a subsequence of the chair sequence by St\"uckler and Behnke. A limitation of our method is that we can process sequences with up to about $200$ frames. We can only segment objects that move within processed sequences, hence we do not separate the chair on the left, as in the ground truth. Figure~\ref{fig:bonn_chair_reconstruction} \red{middle column} shows the accumulated point cloud and reconstructions of selected objects of this sequence. Despite the oversegmentation, the alignments and the reconstruction are quite good. The discontinuity in the back wall is correct; together with the poor geometry of the green segment this makes this scene prone to oversegmentation by our method. 

\begin{figure*}[t]%
%
\begin{center}
\includegraphics[width=\textwidth]{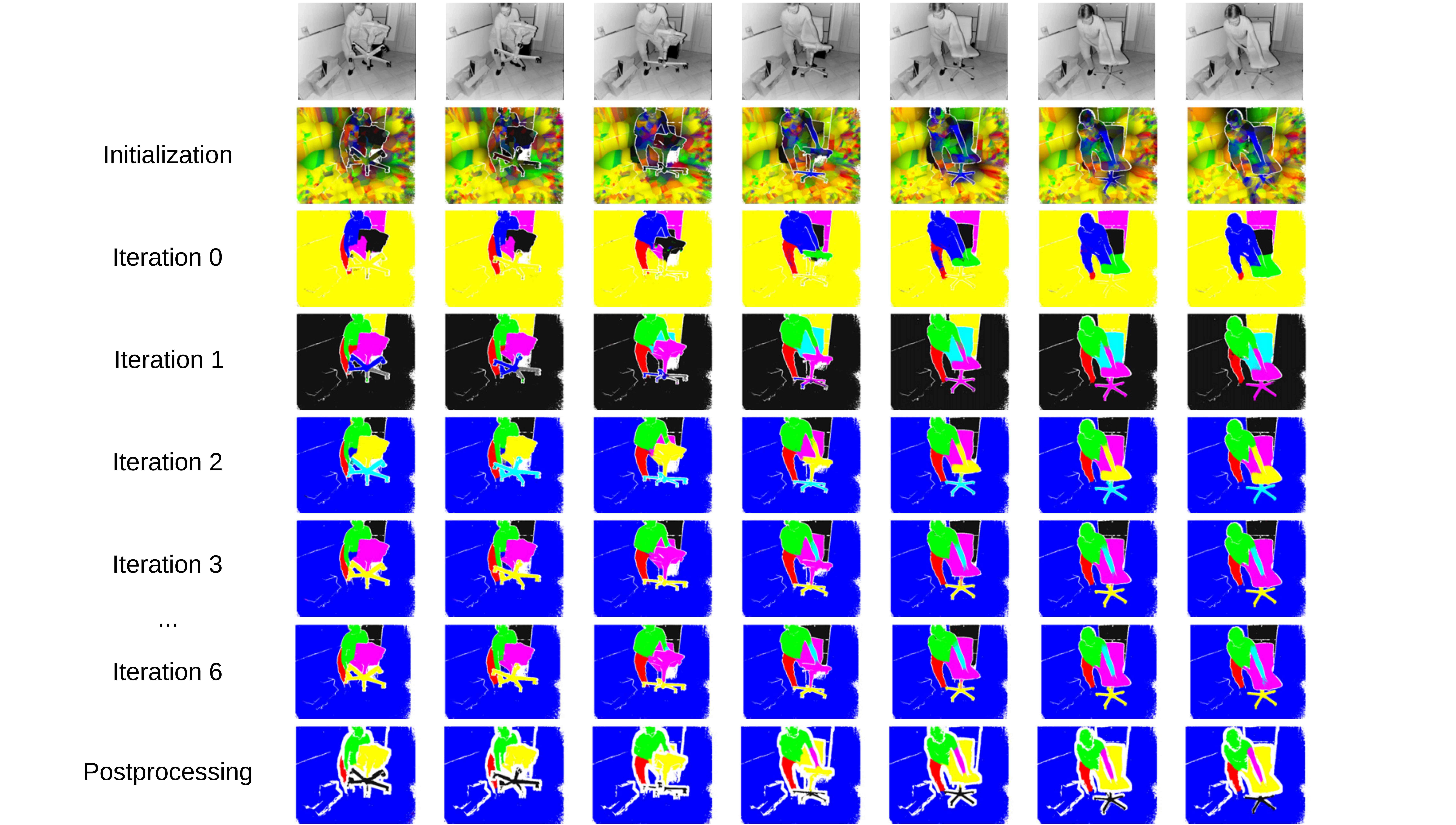}%
\end{center}
\caption{This figure shows our segmentation results throughout the iterations of our optimization. After three iterations the algorithm converges. The temporal consistency and spatial correctness improve steadily (chair seat from iteration 0 to iteration 1, chair wheels from iteration 1 to iteration 2) and the method is robust to the presence of nonrigid objects (body, arms, legs). }%
\label{fig:own_chair_segmentation_convergence}%
\end{figure*}

Figure~\ref{fig:own_chair_segmentation_convergence} shows results from a sequence that we captured using a KinectOne sensor at $20$ frames per second. We used every second frame and a total of $50$ frames for this example. Instead of RGB data we used the infrared data for optical flow during the initialization step (Section~\ref{sec:initialization}), as depth and infrared frames are always perfectly aligned. The sequence involves more complex motion and also a non-rigid person. The person demonstrates the capabilities of the chair by lifting it, spinning the bottom, and putting the chair down again. We are able to consistently segment the bottom from the chair. The non-rigid person is also reasonably segmented into mostly rigid pieces. We also show volumetric reconstructions of selected objects from this sequence in Figure~\ref{fig:ownchair_reconstructions}.

\begin{figure}[t]%
\includegraphics[width=\columnwidth]{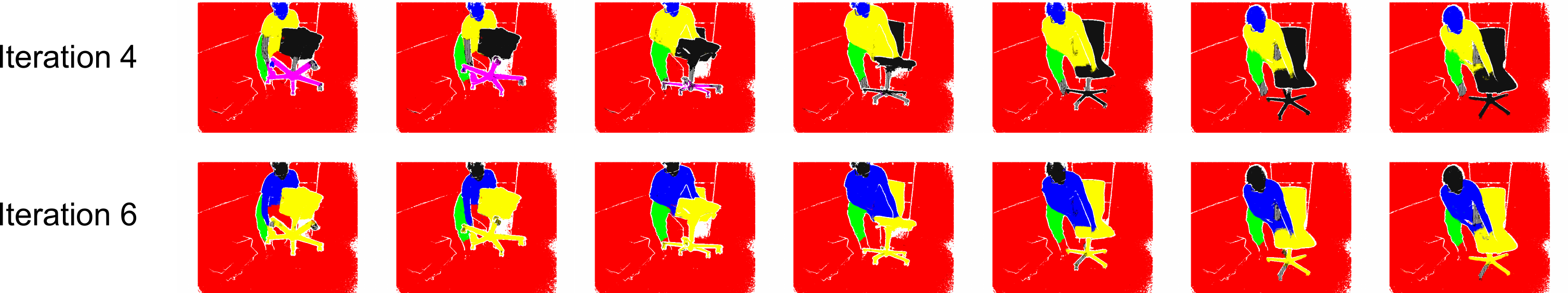}%
\caption{Two iterations of our approach using simple ICP alignments instead of the more complex approach including loop closures (Section~\ref{sec:bundleadjustment}). We do not converge to the desired solution since the ICP alignments are not precise enough.}%
\label{fig:simple_icp_fails}%
\end{figure}

In Figure~\ref{fig:simple_icp_fails} we show two iterations on our chair sequence (Figure~\ref{fig:own_chair_segmentation_convergence}) where simple incremental point to plane ICP is used to find the motion sets (Section~\ref{sec:bundleadjustment}), instead of using our more refined method including loop closures. The bottom of the chair fails to be fused to one segment because the simple ICP method looses track in the frames where the label shift is located.

\begin{figure}%
\includegraphics[width=\columnwidth]{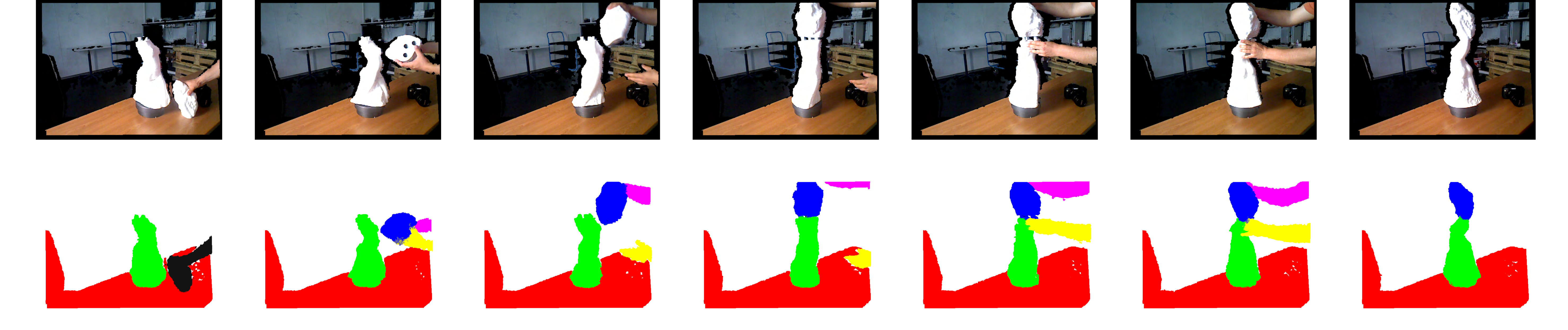}%
\caption{Results from a sequence captured with an Asus Xtion Pro sensor. It contains $220$ frames. The statue is largely textureless. Since our approach mostly relies on geometry, we still obtain good results.}%
\label{fig:statue_sequence}%
\end{figure}

Figure~\ref{fig:statue_sequence} shows results from a sequence that we recorded with an Asus Xtion Pro sensor. Due to high noise levels we ignored all data further than two meters away. The sequence features a rotating statue which is assembled by hand. Our optimization finds the correct segmentation with exception of the statue's head in the beginning. This is due to strong occlusions between the two hands and the statue. To compute these results we used $220$ frames, sampled at ten frames per second.

\section{Conclusions}

We presented a novel method for temporally consistent motion segmentation from RGB-D videos. Our approach is based almost entirely on geometric information, which is advantageous in scenes with little texture or strong appearance changes. We demonstrated successful results on scenes with complex motion, where object parts sometimes move in parallel over parts of the sequences, and their motion trajectories may split or merge at any time. Even in these challenging scenarios we obtain consistent labelings over the entire sequences, thanks to a global energy minimization over all input frames. Our approach includes two key technical contributions: first, a novel initialization approach that is based on clustering sparse point trajectories obtained using optical flow, by exploiting the 3D information in RGB-D data for clustering. Second, we introduce a strategy to generate new object labels. This enables our energy minimization to escape situations where it may be stuck with temporally inconsistent segmentations.

A main limitation of our approach is that due to the global nature of the energy minimization, the length of input sequences that can be processed is limited. In the future, we plan to develop a hierarchical scheme that is able to consistently merge shorter subsequences, which are processed separately in an initial step, of arbitrarily long inputs. Another limitation of our approach is the piecewise rigid motion model, which we also would like to address in the future. Finally, the processing times of our current implementation could be reduced significantly by moving the computations to the GPU.



{\small
\bibliographystyle{ieee}
\bibliography{egbib}
}

\end{document}